\documentclass[runningheads]{llncs}

\usepackage{graphicx}
\usepackage{microtype}
\usepackage[caption=false]{subfig} %side by side images https://tex.stackexchange.com/questions/499870/lecture-notes-in-computer-science-lncs-side-by-side-figures

\usepackage{makecell}  %line breaks in cell  https://tex.stackexchange.com/questions/2441/how-to-add-a-forced-line-break-inside-a-table-cell
\usepackage{xcolor,colortbl}

\definecolor{Gray}{gray}{0.85}
\definecolor{LightCyan}{rgb}{0.88,1,1}
\newcolumntype{a}{>{\columncolor{Gray}}l}
\newcolumntype{b}{>{\columncolor{LightCyan}}l}

\newcolumntype{T}{>{\tiny}c} % define a new column type for \tiny https://tex.stackexchange.com/questions/125182/how-to-change-the-font-size-only-in-one-column-of-a-table-tabular

\usepackage{booktabs} % for professional tables

\usepackage{amsmath}
\usepackage{microtype}
\usepackage{multirow}

\usepackage[acronym]{glossaries}
\newacronym{TOPIC}{TOPIC}{Credit card fraud detection - Classifier selection strategy}
\begin{document}

\title{Credit card fraud detection - Classifier selection strategy}
\titlerunning{Credit card fraud detection - Classifier selection strategy}

%\iffalse %DOUBLE BLIND
\author{Gayan K. Kulatilleke}
\authorrunning{G. K. Kulatilleke}
\institute{Work outlined in this paper is part of the author’s MSc dissertation at Queen Mary University of London, 2017.\\\email{tidalbobo@gmail.com}}
%
%\fi
\maketitle  
\begin{abstract}
Machine learning has opened up new tools for financial fraud detection. Using a sample of annotated transactions, a machine learning classification algorithm learns to detect frauds. With growing credit card transaction volumes and rising fraud percentages there is growing interest in finding appropriate machine learning classifiers for detection.
However, fraud data sets are diverse and exhibit inconsistent characteristics. As a result, a model effective on a given data set is not guaranteed to perform on another. Further, the possibility of temporal drift in data patterns and characteristics over time is high. Additionally, fraud data has massive and varying imbalance. 
In this work, we evaluate sampling methods as a viable pre-processing mechanism to handle imbalance and propose a data-driven classifier selection strategy for characteristic highly imbalanced fraud detection data sets. The model derived based on our selection strategy surpasses peer models, whilst working in more realistic conditions, establishing the effectiveness of the strategy.

\end{abstract}

\section{Introduction} \label{Introduction}

Credit cards play a significant role in our economy \cite{zojaji2016survey}. Unfortunately credit card fraud losses are high and increasing; £618.0 million in 2016, a 9\% increase from the previous year and the fifth consecutive year of increase \footnote{Financial Fraud Action UK (FFA UK)}. Also, prevention rates are low; only 0.6/£ \cite{kulatilleke2022Challenges}. There is also high remedial costs; \$2.40 for each 1\$ fraud, in 2016, an 8\% increase from previous year \footnote{Lexisnexis.com,2018}.

While frauds are around 0.2\% of all card transactions, their large transaction values can be high \cite{sohony2018ensemble}. Thus, fraud detection systems are essential \cite{zareapoor2015application,kulatilleke2022Challenges,kulatilleke2022Empirical}. Fraud detection identifies if a new authorized transaction belongs to the class of fraud or normal transactions \cite{maes2002credit}. Such a process should be economical \cite{quah2008real}, a key factor to consider with growing transaction volume. 

The Credit Card Fraud Detection is one that can be ideally explored in the machine learning, Artificial intelligence, and big data domains \cite{kulatilleke2022Empirical}. This is a sharp contrast to the traditional rule (or exception) based mechanisms and filters that must be painstakingly setup only to be discarded when the fraudsters change their modus-operandi. 

Given a detailed and large labelled sample of historical transactions, machine learning can be used to \textit{learn} complex characteristics and flag frauds in new transactions \cite{kulatilleke2022Challenges}. It has the ability to study large amounts of data to discover hidden as well as evolving patterns of malicious occurrences \cite{sathyapriya2017big}.

However, according to \cite{kulatilleke2022Challenges,kulatilleke2022Empirical} there are wide variations in the fraud detection data set in terms of feature definitions, size, feature size and fraud ratios; however they are all massively unbalanced. This variation in data distributions could explain the inability to obtain an ideal general algorithm. Furthermore, there is no ideal algorithm known in credit card fraud literature that outperforms all others \cite{zojaji2016survey,kulatilleke2022Empirical}. 

The ability to build an accurate predictive model for characteristic massively imbalanced, possibly encoded data sets has implications in modelling bank and firm collapse, nuclear plant/aircraft failure and similar rare occurring yet not-to-be missed events \cite{kulatilleke2022Empirical}.

In this work, We propose a data driven testing and assessing approach to dynamically select an appropriate model, based on the score of the evaluation metric and balancing strategy. 
Specifically we: 
\begin{itemize}
    \item Investigate the effects of balancing strategies and algorithms on performance of classifiers
    \item compere the computational training time cost of a wide range of models across multiple balancing strategies
    \item Propose a universally effective classifier selection strategy for deriving an accurate model from any given massively unbalanced data set without any prior information
\end{itemize}

\section{Background}

\begin{table}[ht]
    \scriptsize
    \begin{tabular}{l} \\ \toprule
    %Acronyms                                                 \\ \midrule
    Ada\_Disc - AdaBoost Classifier using discrete boosting algorithm \\
    Ada\_Real - AdaBoost Classifier using real boosting algorithm     \\
    AuROC - Area under ROC                                           \\
    DT - Decision Tree Classifier                                    \\
    FN - false negatives                                             \\
    FP - false positives                                             \\
    FPR - False Positive Rate                                        \\
    GA - Genetic Algorithem                                          \\
    GBC - Gradient Boosting Classifier                               \\
    GNB - Gaussian Naive Bays                                        \\
    KNN - KNeighbors Classifier                                       \\
    LR - Logistic Regression                                         \\
    NN\_MLP - MLP Classifier                                           \\
    PAClassifier - Passive Aggressive Classifier                     \\
    PCA - Principle Component Analysis                               \\
    QuadraticDA - Quadratic Discriminant Analysis                    \\
    RF - Random Forest Classifier                                     \\
    Ridge - Ridge Classifier                                          \\
    ROI - Return On Investment                                       \\
    SGD - SGD Classifier                                             \\
    SVC\_Linear SVC using the linear kernel                           \\
    SVM - Support Vector Machine                                     \\
    TN - true negatives                                              \\
    TNR - True Negative Rate                                         \\
    TP - true positives                                              \\
    TPR - True Positive Rate                                         \\ \bottomrule              
    \end{tabular}
    \caption{Machine learning models and evaluation metrics acronyms used.}
    \label{table-acronyms}
\end{table}
Table~\ref{table-acronyms} lists the acronyms and terminology we use, particularly in the charts. 

\subsection{Machine learning}
Machine learning techniques available for fraud detection can be separated as supervised, semi-supervised and unsupervised \cite{buczak2015survey}. Supervised learning uses labelled fraud and normal transactions to learn a model that is able to identify frauds in new transactions (in case of classification) or provide a risk rating (in case of regression) so that investigators can prioritize on a subset of highly probable frauds. Unsupervised learning attempts to cluster transactions into fraud and normal based on similarities in features and does not require labelled data \cite{sathyapriya2017big}. Semi-supervised techniques use partly labelled data \cite{pise2008survey}. Many machine learning techniques such as NN can be used in all modes, though supervised algorithms and models are most common \cite{duman2013novel}.

Some researchers \cite{zojaji2016survey} consider behavior analysis as a valid approach for fraud detection. The assumption is that a frustrater's use of the card would deviate from the authorized user’s behavior and thus could be detected as an anomaly as it stands out. This is also known in literature as a one-class classification \cite{zheng2019one}. It has the benefit of only requiring the normal data for the learning process, which is more common and consists of nearly 99\% of the transaction volume \cite{duman2013solving}. It does however require a model that understands a ‘normal’ user behavior \cite{zojaji2016survey}. However, while user behavior analysis is able to detect novel frauds easily \cite{zojaji2016survey} that classifiers may miss, it has a high rate of false alarms (FP) that, given today’s huge transaction rates makes these unattractive \cite{zojaji2016survey}. Also, \cite{zheng2019one} reports that sophisticated fraudsters deliberately emulate normal user patterns, making void the different-behavior assumption which is the basis of these models. According to \cite{zheng2019one}, one-class classifiers also suffer from the need for a domain expert to set the detection threshold manually.

Unsupervised models require excessive training and time to reach acceptable levels of performance \cite{zojaji2016survey}, which becomes impractical due to the rapidly evolving nature of frauds over short time spans. As our work is based on a PCA encoded data set, it is not possible to use any semi-supervised approaches.

Therefore, all models considered in this work are based on supervised learning algorithms which uses human labelled fraud and normal transactions to create a predictive model \cite{zojaji2016survey}. After the model is developed and trained, its "learnings" can be stored, duplicated and used on multiple and different systems \cite{buczak2015survey} and consists of the distilled knowledge which is simply a set of numbers that is fraction of the training data set size.

\subsection{Model options}
According to \cite{sahin2011detecting,duman2013solving}, NN, DT, SVM, LR and KNN are among the popular detection methods. They can be used independently or as part of an ensemble which combines several models by voting or averaging.

NN are non-linear learning models that can be made arbitrary complex by adding layers of neurons. It mimics the connected neural structure of the human brain, with the neuron being the basic element. A neuron has the ability to learn a non-linear transfer function. Generally, there is input, hidden and output layers of neurons, interconnected to the to the preceding layer with weights associated that determines the level of influence. Back propagation is used to adjust the weights so that NN is able to minimize predictive error and achieve learning. \cite{sahin2011detecting} reports that NN outperforms LR, though this is based on Accuracy score, and acknowledges this as a limitation. However, \cite{dal2017credit} states that several studies have reported RF to achieve the best performance.

While DT works on balanced data sets and is known to be good at poorly balanced data set, \cite{duman2013solving} noted its poor performance due to lesser number of leaf nodes being created, when the frauds to normal is massively imbalanced. However, a DT is easy to understand and can be human interpreted \cite{zojaji2016survey}.

SVM is a binary classifier that attempts to find a hyper-plane to separate the frauds and normals linearly. While in a lower dimensional space this is not possible, SVM uses what is known as a kernel function to elevate the data into a suitable higher dimensional space, where linear separation becomes possible \cite{zojaji2016survey}. The hyper plane is located between the support vectors and the entire learning process is to identify these few support vectors, which means that the model can perform very fast and with minimal resources once trained. \cite{zojaji2016survey} notes that Radial basis function (RBF) Kernel could learn complex input spaces. Yet, DT is known to outperforms SVM in credit card fraud detection \cite{yeh2009comparisons}.

In addition, \cite{zheng2019one} has used a modified GAN (Generative adversarial networks) based generative model with one-class classification (OCAN, one-class adversarial net) for fraud detection. In their paper they have published the results for the same data set that this work uses. Therefore, the GAN model is not repeated. As OCAN outperforms other GAN implementations \cite{zheng2019one}, we will compare its results with ours, later, to asses relative effectiveness.

Interestingly, ensembles combine predictions of many models with the aim of improving generalizability and robustness over a single model. There are 2 main ensemble methods in the scikit-learn python library \footnote{Scikit-learn.org, 2018}.

\begin{itemize}
    \item Averaging methods - use several independent models and averages their predictions. As averaging reduces variance, the combined estimator will be better than its individual components. Examples consists of Bagging methods, Forests of randomized trees etc.
    \item Boosting methods - uses weak base models cascaded sequentially to construct a more powerful model. 
    Examples are AdaBoost, Gradient Tree Boosting etc.
\end{itemize}

Despite consisting of a collection of classifiers, ensembles are generally fast and efficient. \cite{dutta2016ensemble} reports that their multi-class ensemble learner beat an SVM model comfortably. Therefore, this is a promising area for experimentation. 

GA \cite{holland1992genetic} mimics evolution and attempts to evolve a best-fit solution using iteration and mutation. The strength of a solution is measured by its ability to solve the underlying problem, which is scored by its fitness. Each new generation should have a better fitness, and once fitness stops improving, the learning is complete. While it is a popular and strong algorithm in credit card fraud detection \cite{zojaji2016survey}, we do not consider it due to the extensive CPU and memory constraints.

Many popular methods such as LR, KNN and NN exceeds the practical time limits available \cite{zojaji2016survey}. One way past this hurdle is the use of PCA which reduces the dimensionality of the data.

\subsection{Unbalanced classification}

Fraud detection problem is difficult due to massive imbalance \cite{duman2013solving}.
Further, these minute fraud fractions suffer from concept drift, where transactions might change their statistical properties over time \cite{dal2015calibrating}, often from perpetrators adapting different tactics over time \cite{kulatilleke2022Challenges}.

As the data is massively unbalanced, to enable the model to learn both the (minority) frauds and normal (majority), some form of balancing (undersampling or oversampling) should be performed \cite{sahin2011detecting,kulatilleke2022Challenges}. Otherwise, the imbalance severely interferes with the learning ability of most algorithms and produces unreliable models \cite{sohony2018ensemble}. Specifically, \cite{west2016some} observed genetic algorithms, neural networks, SVM and fuzzy logic all had significantly inferior specificity scores attributed to overfitting due to massive imbalance.

Sampling either lowers the majority class (undersampling) or raise the minority (oversampling) to re-balance the data \cite{kulatilleke2022Challenges}. Balancing has a direct influence on the resultant model \cite{dal2015calibrating}. While oversampling replicates the minority class providing the classifier more opportunity to learn, there is no new information, and may lead to overfitting the majority class especially in the case of noisy input \cite{sohony2018ensemble}. Further, oversampling increases training time as it creates new copies \cite{chawla2002smote}.

Undersampling will discard the majority class \cite{kulatilleke2022Challenges}. While it leads to a significant loss of data, in contrast to oversampling, it creates a model on real observed data. However, undersampling still alters the priors which consequently biases the posterior probabilities of a classifier \cite{dal2015calibrating}. 

Both oversampling and undersampling alters the training data set distribution from that of the real data set, leading to a biased training set, referred to as the sample selection bias \cite{kulatilleke2022Challenges}. A common solution for sample selection bias is using importance weighting or ensemble classifiers \cite{dal2017credit}. 

\begin{table}[]
    \begin{tabular}{p{0.70\linewidth} | >{\centering\arraybackslash}p{0.30\textwidth} }
    \toprule
    Undersampling method                                                    & Reference             \\ \midrule
    Random majority under-sampling with   replacement                       &                       \\
    Instance Hardness Threshold                                             & Smith et. al. (2014) \cite{smith2014instance}  \\ \bottomrule
    \end{tabular}
    \caption{Undersampling methods in the in the imbalanced-learn python library}
    \label{TABLE_undersampleing}
\end{table}
\begin{table}[]
    \begin{tabular}{p{0.70\linewidth} | >{\centering\arraybackslash}p{0.30\textwidth} }
    \toprule
    Oversampling method                                                     & Reference             \\ \midrule
    Random minority over-sampling with   replacement                        &                       \\
    SMOTE - Synthetic Minority Over-sampling   Technique                    & Chawla et. al. (2002) \cite{chawla2002smote} \\
    ADASYN - Adaptive synthetic sampling approach for imbalanced learning & He et al. (2008) \cite{he2008adasyn}    \\ \bottomrule
    \end{tabular}
    \caption{Oversampling methods in the in the imbalanced-learn python library}
    \label{TABLE_oversampleing}
\end{table}

Table~\ref{TABLE_undersampleing} shows some of the undersampling methods implemented in the imbalanced-learn python library \footnote{Contrib.scikit-learn.org, 2018} which is used later in this work for experiments and classifier selection. The imbalanced-learn python library also provides oversampling methods, which are given in Table~\ref{TABLE_oversampleing}. 

There is also more complex "oversampling followed by undersampling" implementations such as SMOTE + Tomek links \cite{batista2003balancing}, SMOTE + ENN \cite{batista2004study} as well as Ensemble samplings EasyEnsemble \cite{liu2008exploratory} and BalanceCascade \cite{liu2008exploratory}.

\subsection{Typical fraud detection data characteristics}
There is a huge diversity and conflicting characteristics (feature definitions, depth and breadth, fraud ratios) in fraud data sets \cite{kulatilleke2022Challenges}. The lack of a corpus that defines what should be typical credit card data, despite the universal and ubiquitous nature of card payments \cite{kulatilleke2022Empirical}, is another significant challenge. Given the only similarity amongst fraud detection data is the characteristic massively unbalanced nature \cite{kulatilleke2022Challenges}, a data-driven approach for model selection needs to be adopted.

\section{Methodology}
\subsection{Framework for massively unbalanced unknown data sets}

\begin{figure}
    \centering \includegraphics[width=0.8\columnwidth]{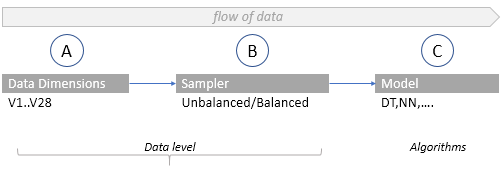}
    \caption{Data flow model and activity intervention nodes}
    \label{fig_frameworkDataflow}
\end{figure}

Figure~\ref{fig_frameworkDataflow} shows data transitioning from source to the final classifier through 3 intervention nodes A, B and C. Node A performs dimensional reduction. 
When using PCA data, as explained later, its principle components V1 to V28 successfully declines in significance. Node B facilitates some form of balancing using any sampling method. We investigat 2 undersampling methods (Random with replacement, Instance Hardness Threshold) and 3 oversampling methods (Random oversampling, SMOTE, ADASYN) from the imbalanced-learn library. Finally, Node C incorporates a classifier model.

This work experimented with 15 models from the scikik-learn library which includes DummyClassifier, LogisticRegression, RandomForestClassifier, GaussianNB, SVC(Linear), MLPClassifier, RidgeClassifier, DecisionTreeClassifier, SGDClassifier, PassiveAggressiveClassifier, Perceptron, KneighborsClassifier and ensembles AdaBoostClassifier (Real), AdaBoostClassifier (Discrete), GradientBoostingClassifier along with QuadraticDiscriminantAnalysis. A total set of 28 x 15 unbalanced and 5 x 15 balanced combinations of models were studied using the Colaborotory cloud-based environment. A typical run on needed about 5-6 hours to complete with the full primary data set. 

\begin{figure}
    \centering \includegraphics[width=0.8\columnwidth]{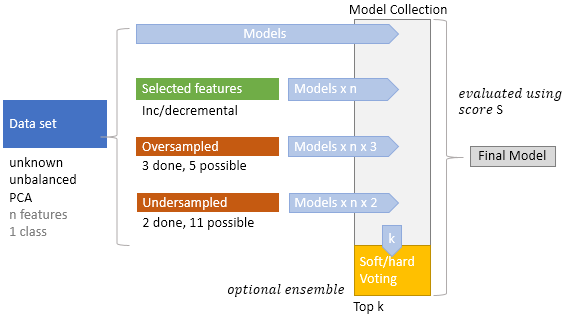}
    \caption{Logical diagram for the classifier selection framework}
    \label{fig_frameworkLogical}
\end{figure}
Figure~\ref{fig_frameworkLogical} shows the logical diagram of the proposed framework. It assumes an unknown massively unbalanced data set is given as input, creates a collection of models based on the A-B-C node permutations, evaluates these on a given metric $S$ and selects the best scoring $k$ set of classifiers. Sample Selection Bias can be compensated using the ensemble-based implementation at the final stage which is an optional refinement.

Each model uses stratified cross-validation (i.e.: class proportion in the data sets is kept constant for all folds) and a 20:80 test: train split. Balancing is only for the training set, to avoid information bleeding from validation to training data and maintain test data characteristics identical to the original. Multiple evaluation metrics including F1 and g-mean is calculated in $S$ allowing different metrics to be used based on the objective.

\subsection{Data sets}
Researchers need to make a trade-off between the real world and practical data, due to constraints such as privacy and confidentiality, especially in domains as credit card transactions. It was shown that PCA transformation is a suitable obfuscation method that does not damage information content significantly \cite{kulatilleke2022Challenges}. 
Thus, we use 2 publicly available data sets in PCA encoded format. 

The primary data set \cite{dal2015calibrating} is a real-world credit card transaction data set released to public domain by the ULB Machine Learning Group (MLG), a research unit of the Computer Science Department of the Faculty of Sciences, Université libre de Bruxelles. It has 284,807 annotated  transactions related to 2 days in September 2013 by European cardholders. There are 492 or 0.172\% frauds, indicating massively unbalanced data. Features are encoded in the 28 principle components $V1 \dots V28$. Due to PCA encoding, it is not possible to obtain any semantic information from the features, and it is essentially obfuscated, with the exception of the transaction date and amount (which are the 2 raw features) as shown in Table~\ref{TABLE_PCA_primary_dataset}. 
\begin{table}[t]
    \begin{tabular}{p{0.20\linewidth} p{0.20\linewidth} p{0.40\linewidth} p{0.20\linewidth}}
    \toprule
    Normal  & Frauds & Features            & Instances \\ \midrule
    284,315 & 492    & 30 (28 PCA + 2 Raw) & 284, 807  \\ \bottomrule
    \end{tabular}
    \caption{Primary PCA encoded (Pozzolo) Data set }
    \label{TABLE_PCA_primary_dataset}
\end{table}

Our secondary data set, summarized in Table~\ref{TABLE_secondary_data set}, was used by \cite{yeh2009comparisons} for binary classifier evaluation. It is publicly available from the University of California Irvine (UCI) Machine learning repository \footnote{Archive.ics.uci.edu, 2018}. It consists of raw (non PCA encoded) client information, payment history, status and credit limits of credit card customers in Taiwan from April 2005 to September 2005 \cite{lichman2013uci}. For the purpose of our classifier evaluation, we PCA encode this data set, to match our primary data set. Specifically we create 23 principle components using full SVD with a standard LAPACK solver from the scikit library.
\begin{table}[t]
    \begin{tabular}{p{0.20\linewidth} p{0.20\linewidth} p{0.40\linewidth} p{0.20\linewidth}}
    \toprule
    Normal  & Frauds & Features  & Instances \\ \midrule
    23,364  & 6634   & 23 Raw   & 30,000   \\ \bottomrule
    \end{tabular}
    \caption{Secondary Data set }
    \label{TABLE_secondary_data set}
\end{table}

\subsection{Implementation}
Python is used as the main programming language. All code and models were run on Google Colaboratory, a cloud based Jupyter notebook environment, with GPU support and Google Drive based storage.
Scikit-learn and imbalanced-learn are the main libraries used. Imbalanced-learn is a python package offering several re-sampling techniques commonly applied for data sets with massive imbalance. It is compatible with scikit-learn and is part of scikit-learn-contrib projects.

Using the framework, next sections will examine algorithmic efficiencies on unbalanced vs balanced data and efficiencies on dimensional reduction and feature elimination

\section{Results}
\subsection{Primary data set - Algorithmic efficiencies on unbalanced vs Balanced Data}
\begin{table*}[]
    \centering
    \scriptsize
    \setlength{\tabcolsep}{3pt}
    \begin{tabular}{llll ab    llll|r} \toprule
    Model        & Acc & Preci & Recall & F1     & G-mean & auroc  & Cohen  & Matthew & Hamm & Time(s)  \\ \midrule
    KNN          & 0.9994   & 0.9114    & 0.7347 & 0.8136 & 0.8571 & 0.8673 & 0.8133 & 0.818   & 0.0006  & 486.6543 \\
    RF           & 0.9994   & 0.9211    & 0.7143 & 0.8046 & 0.8451 & 0.8571 & 0.8043 & 0.8108  & 0.0006  & 18.8989  \\
    SVC-Linear   & 0.9993   & 0.8315    & 0.7551 & 0.7914 & 0.8689 & 0.8774 & 0.7911 & 0.792   & 0.0007  & 701.3156 \\
    DT           & 0.9992   & 0.7684    & 0.7449 & 0.7565 & 0.8629 & 0.8723 & 0.7561 & 0.7562  & 0.0008  & 20.9145  \\
    Perceptron   & 0.9991   & 0.7143    & 0.7653 & 0.7389 & 0.8746 & 0.8824 & 0.7385 & 0.7389  & 0.0009  & 0.9337   \\
    PAClassifier & 0.9991   & 0.73      & 0.7449 & 0.7374 & 0.8629 & 0.8722 & 0.7369 & 0.737   & 0.0009  & 0.9703   \\
    Ada-Real     & 0.9991   & 0.7857    & 0.6735 & 0.7253 & 0.8205 & 0.8366 & 0.7248 & 0.727   & 0.0009  & 60.2403  \\
    Ada-Disc     & 0.9991   & 0.7647    & 0.6633 & 0.7104 & 0.8143 & 0.8315 & 0.7099 & 0.7117  & 0.0009  & 58.4849  \\
    GBC          & 0.9991   & 0.7922    & 0.6224 & 0.6971 & 0.7888 & 0.8111 & 0.6967 & 0.7018  & 0.0009  & 93.7396  \\
    LR           & 0.9991   & 0.8889    & 0.5714 & 0.6957 & 0.7559 & 0.7857 & 0.6952 & 0.7123  & 0.0009  & 3.4274   \\
    NN-MLP       & 0.999    & 0.8475    & 0.5102 & 0.6369 & 0.7142 & 0.755  & 0.6365 & 0.6571  & 0.001   & 12.0008  \\
    SGD          & 0.999    & 0.875     & 0.5    & 0.6364 & 0.7071 & 0.7499 & 0.6359 & 0.661   & 0.001   & 0.9496   \\
    Ridge        & 0.9989   & 0.8367    & 0.4184 & 0.5578 & 0.6468 & 0.7091 & 0.5573 & 0.5912  & 0.0011  & 0.8228   \\
    GNB          & 0.9774   & 0.0607    & 0.8367 & 0.1131 & 0.9045 & 0.9072 & 0.1102 & 0.2217  & 0.0226  & 0.7898   \\
    QuadraticDA  & 0.9758   & 0.0591    & 0.8776 & 0.1108 & 0.9254 & 0.9267 & 0.1079 & 0.2242  & 0.0242  & 1.3282   \\
    Dummy        & 0.0017   & 0.0017    & 1      & 0.0034 & 0      & 0.5    & 0      & 0       & 0.9983  & 0.6318   \\ \bottomrule
    \end{tabular}
    \caption{Primary data set with all unbalanced scores. Acc denotes Accuracy, Preci denotes Precision.}
    \label{TABLE_primary_classifier}
\end{table*}
In Table~\ref{TABLE_primary_classifier}, while KNN gives the best F1 score, the faster RF may be more promising, if the focus is on real time detection. It can be seen that all models perform better than  Dummy (the baseline) set, which is a useful for metrics that evaluate a non-majority class.

\begin{figure}
    \centering \includegraphics[width=0.9\columnwidth]{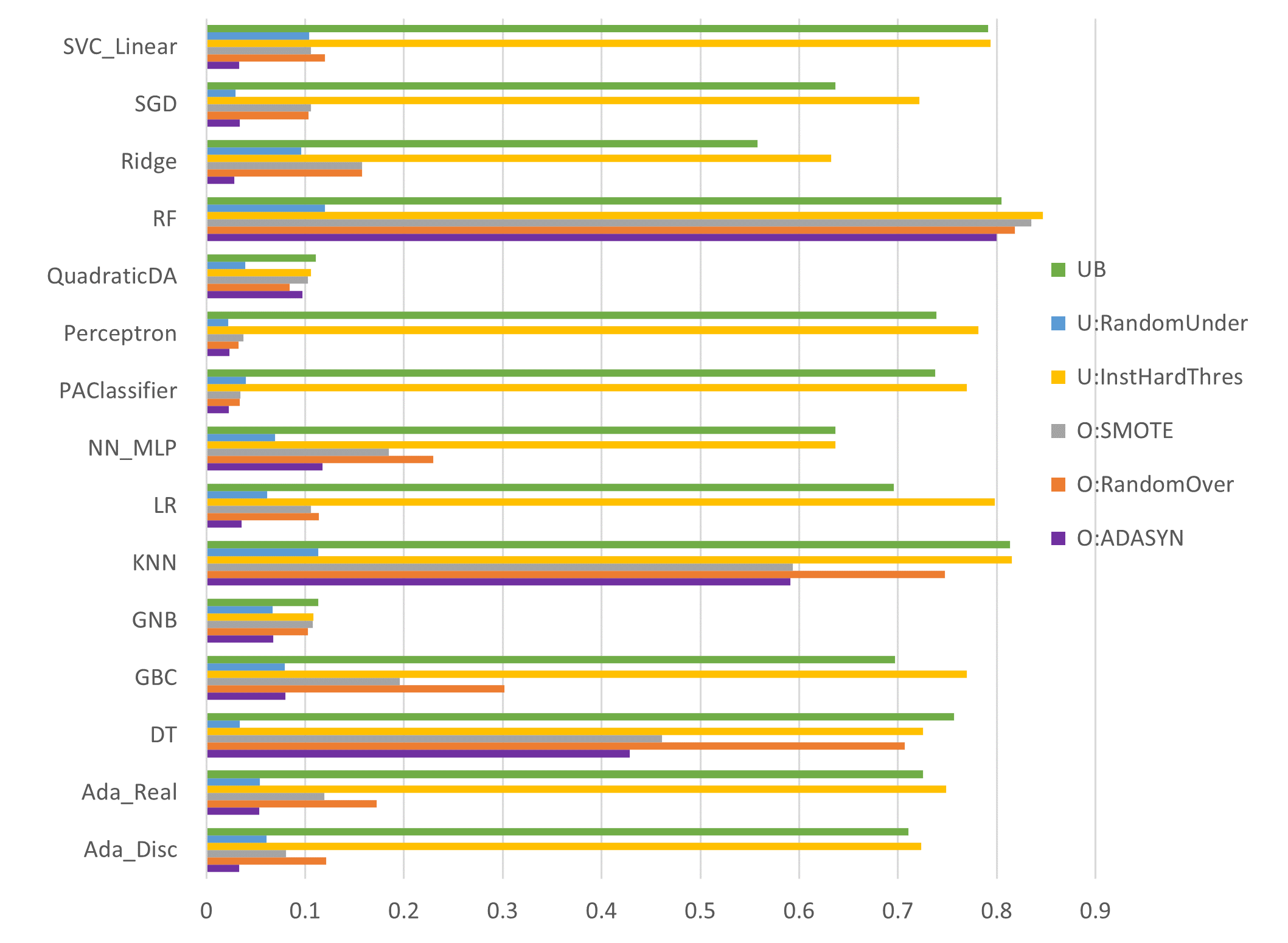}
    \caption{F1 score - unbalanced vs balanced data on the primary data set}
    \label{fig_F1_primary}
\end{figure}

\begin{figure}
    \centering \includegraphics[width=0.9\columnwidth]{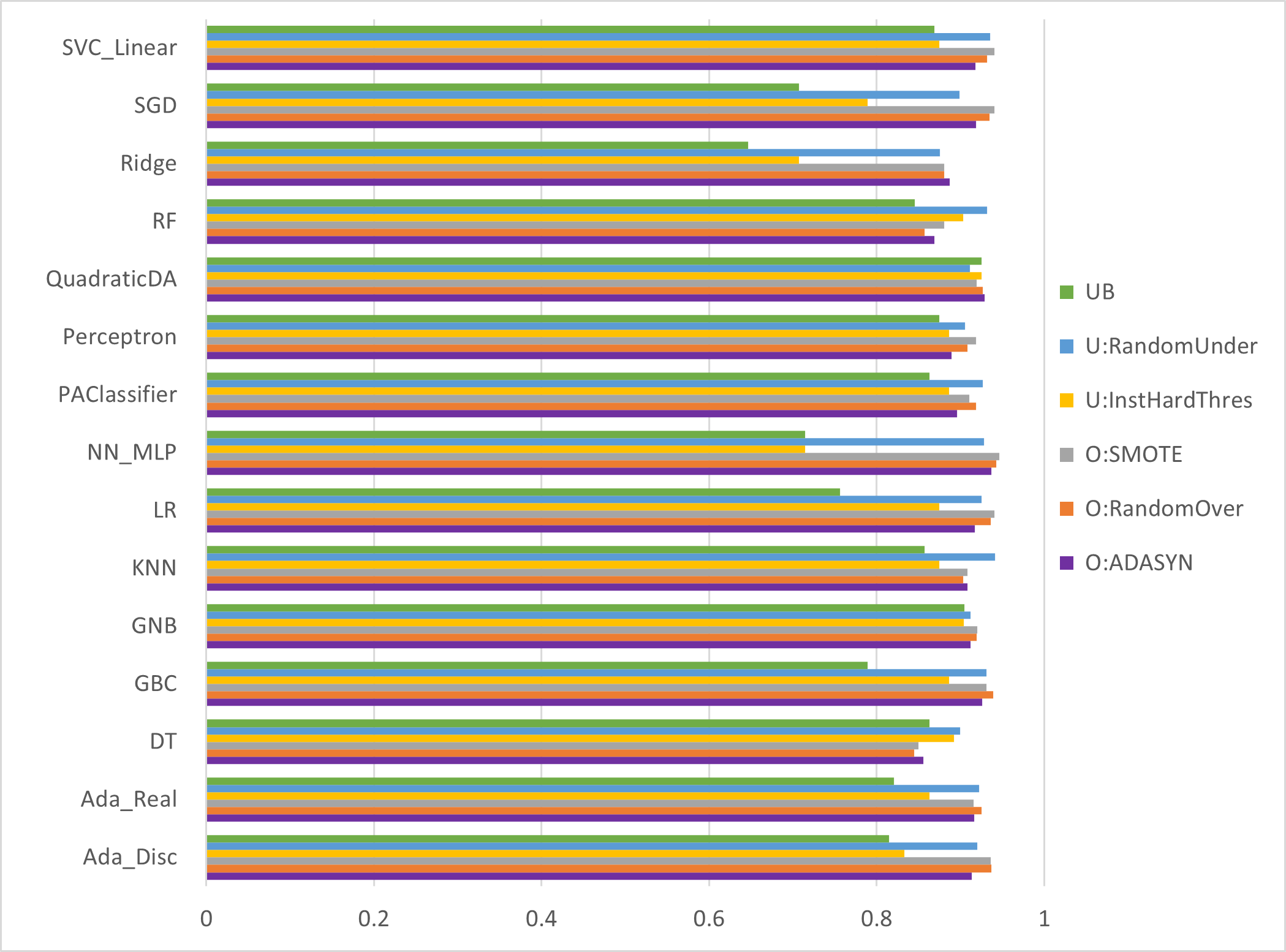}
    \caption{g-mean score - unbalanced vs balanced data on the primary data set}
    \label{fig_Gmeans_primary}
\end{figure}

Figure~\ref{fig_F1_primary} shows the F1 score for all the balancing options against the unbalanced data set. While balancing does not add a significant increase to the F1 score, in order to achieve the extra accuracy for the top scoring models, some form of balancing is important. RF exhibits the best score of 0.8046 unbalanced, which can be increased by 5\% using instance hardness threshold undersampling. However \textit{random} undersampling seems to degrade performance significantly. 

According to the results, undersampling seems to always triumph oversampling. An explanation is that as oversampling adds new interpolated (guessed) fraud data points, which can result in degrading the final classifier.

The g-mean results are shown in Figure~\ref{fig_Gmeans_primary}. 

\subsection{Primary data set - Algorithmic efficiencies on dimensionality reduction}
\begin{figure}[t]
    \centering \includegraphics[width=0.8\columnwidth]{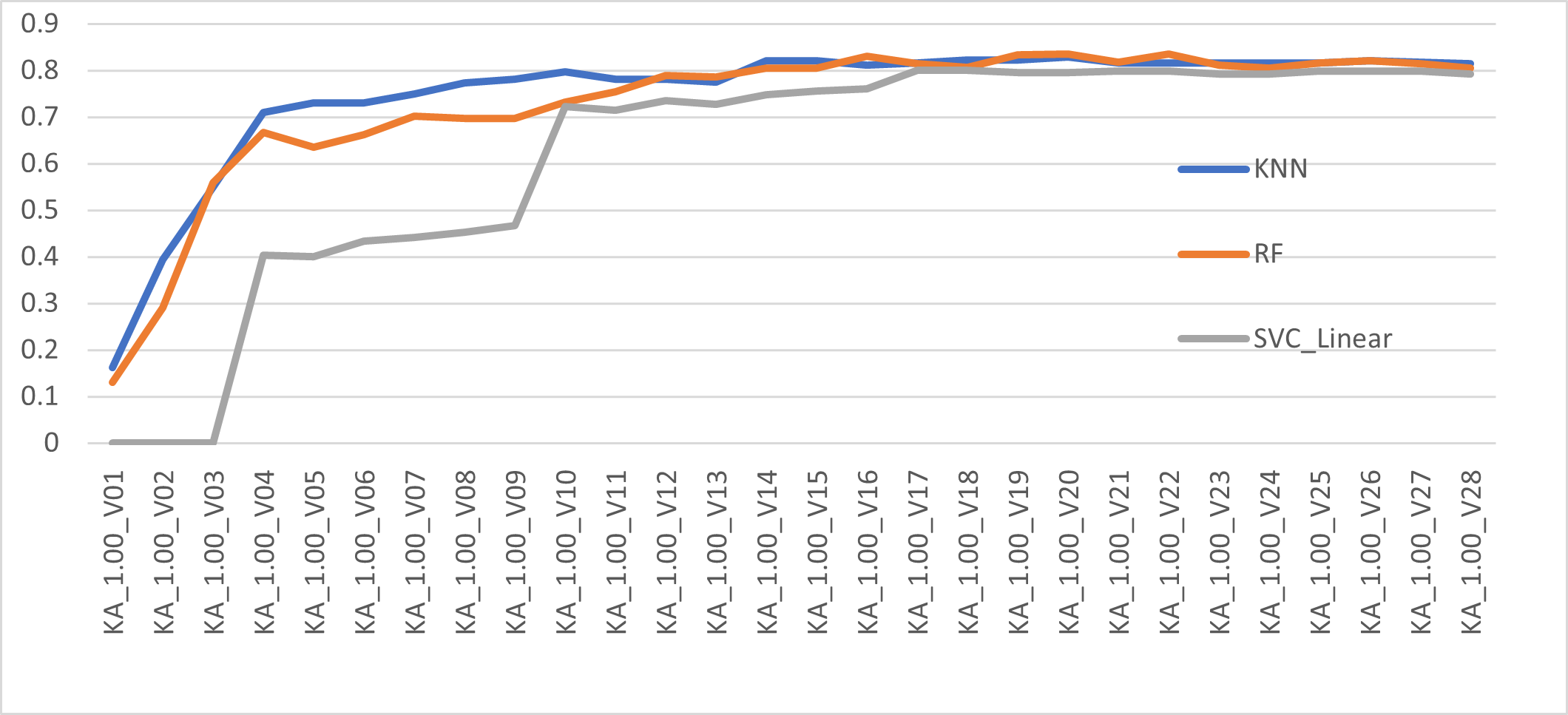}
    \caption{F1 score - unbalanced vs balanced data on the primary data set}
    \label{fig_F1_primaryDIM}
\end{figure}

\begin{figure}
    \centering \includegraphics[width=0.8\columnwidth]{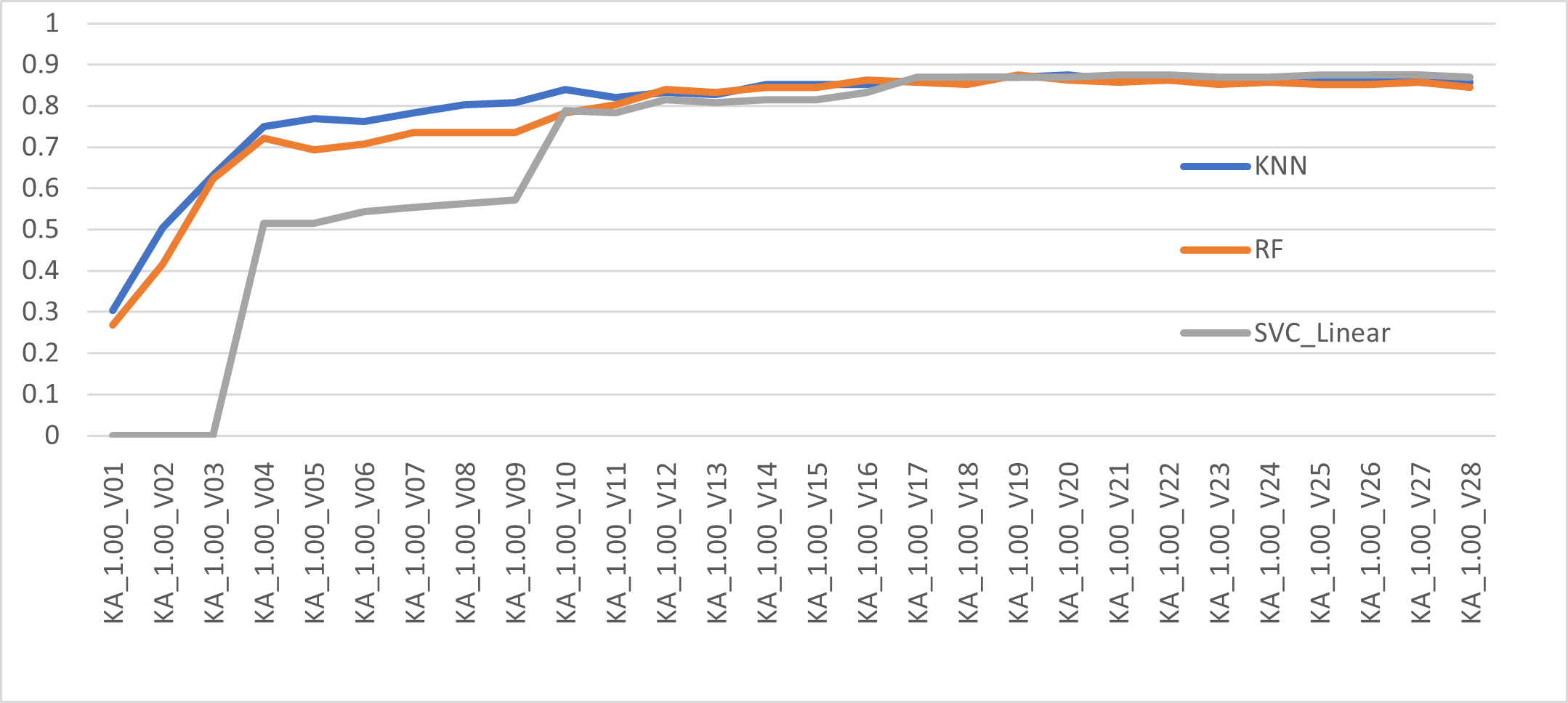}
    \caption{g-mean score - for different dimensions on the primary data set}
    \label{fig_Gmeans_primaryDIM}
\end{figure}

Figure~\ref{fig_F1_primaryDIM} shows the change of F1 score based on the dimensionality. We use full data set with all features.
The top  3 models peek around the first 15 principle components. Also, thereafter, minor degradation on the F1 score is visible. This is most probably due to overfitting in the presence of a strong negative class majority. The g-mean, as seen in Figure~\ref{fig_Gmeans_primaryDIM}, confirms the F1 score findings highlighting the importance of avoiding unwanted features in order to extract the final increase in accuracy.

Therefore, while PCA does not significantly degrade performance, care should be taken to use the appropriate principle component size.

\subsection{Combined results and observations}
Finally, in Table~\ref{TABLE_primary_classifier_ALL}, the combined results for all the 530 combinations and 2 ensembles created using the top 3 (K = 3) F1 score classifiers is presented. RF indicated the best performance.

Interestingly, the soft voting mode (using probability estimates) performed weaker than the hard-voting mode (using summed-up votes). While, in theory, an ensemble should be able to derive a better composite classifier by aggregating the predictions of each component classifier, the experimental results show that this may not always be the case in massively imbalanced data, and an interesting area of future study.

\begin{table*}[]
    \centering
    \scriptsize
    \setlength{\tabcolsep}{3pt}
    \begin{tabular}{llllll ab    lll|r} \toprule
    Model      & S         & Dim     & Acc & Preci & Recall & F1     & G-mean & auroc  & Cohen  & Matthew & Time    \\ \midrule
    RF         & U:I  & 1-28 & 0.9995   & 0.8791    & 0.8163 & 0.8466 & 0.9034 & 0.9081 & 0.8463 & 0.8469  & 134.585 \\
    RF         & O:S          & 1-28 & 0.9995   & 0.9048    & 0.7755 & 0.8352 & 0.8806 & 0.8877 & 0.8349 & 0.8374  & 51.1089 \\
    RF         & UB               & 1-20 & 0.9995   & 0.9481    & 0.7449 & 0.8343 & 0.863  & 0.8724 & 0.834  & 0.8401  & 15.0409 \\
    RF         & UB               & 1-22 & 0.9995   & 0.9481    & 0.7449 & 0.8343 & 0.863  & 0.8724 & 0.834  & 0.8401  & 15.8776 \\
    RF         & UB               & 1-19 & 0.9995   & 0.9146    & 0.7653 & 0.8333 & 0.8748 & 0.8826 & 0.8331 & 0.8364  & 15.5344 \\
    \multicolumn{2}{l}{Vote Hard} & 1-28 & 0.9995   & 0.925     & 0.7551 & 0.8315 & 0.8689 & 0.8775 & 0.8312 & 0.8355  & 231.594 \\
    RF         & UB               & 1-16 & 0.9995   & 0.9359    & 0.7449 & 0.8295 & 0.863  & 0.8724 & 0.8293 & 0.8347  & 14.314  \\
    KNN        & UB               & 1-20 & 0.9995   & 0.9036    & 0.7653 & 0.8287 & 0.8748 & 0.8826 & 0.8285 & 0.8313  & 131.767 \\
    Vote Soft  &                  & 1-28 & 0.9994   & 0.875     & 0.7857 & 0.828  & 0.8863 & 0.8928 & 0.8277 & 0.8289  & 200.917 \\
    KNN        & UB               & 1-18 & 0.9994   & 0.9024    & 0.7551 & 0.8222 & 0.8689 & 0.8775 & 0.8219 & 0.8252  & 98.5816 \\ \bottomrule
    \end{tabular}
    \caption{Primary data set best results. Acc denotes Accuracy, Preci denotes Precision, UB denotes unbalanced, U:I denotes Undersampled-InstHardThres, O:S denotes oversampled-SMOTE and S denotes Sampler.}
    \label{TABLE_primary_classifier_ALL}
\end{table*}

\subsection{Secondary Data set - Experimental Results and Analysis}
\begin{figure}[]
    \centering \includegraphics[width=0.9\columnwidth]{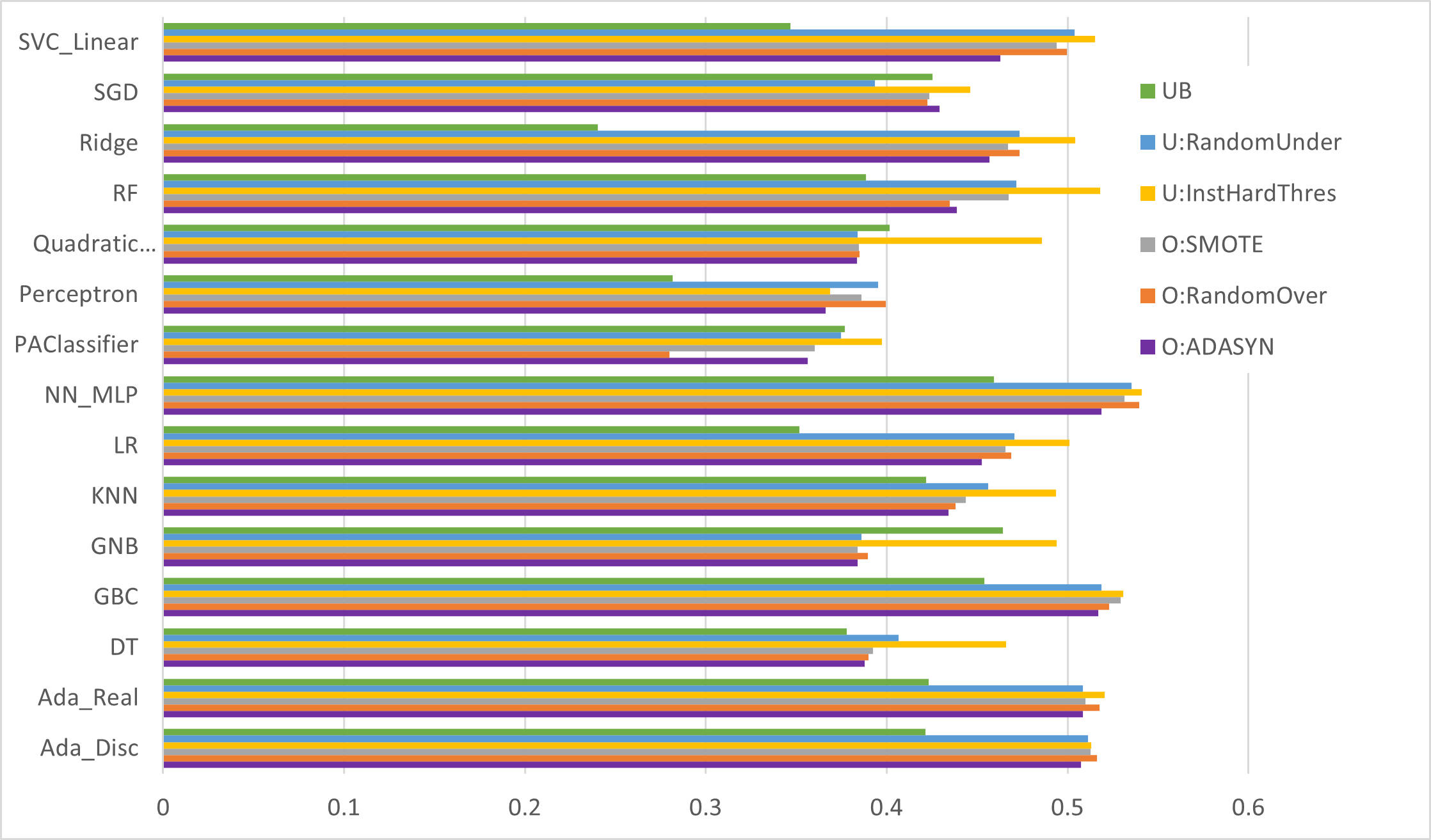}
    \caption{F1 score - Effect of balancing - secondary data set}
    \label{fig_F1_sec}
\end{figure}

\begin{figure}[]
    \centering \includegraphics[width=0.9\columnwidth]{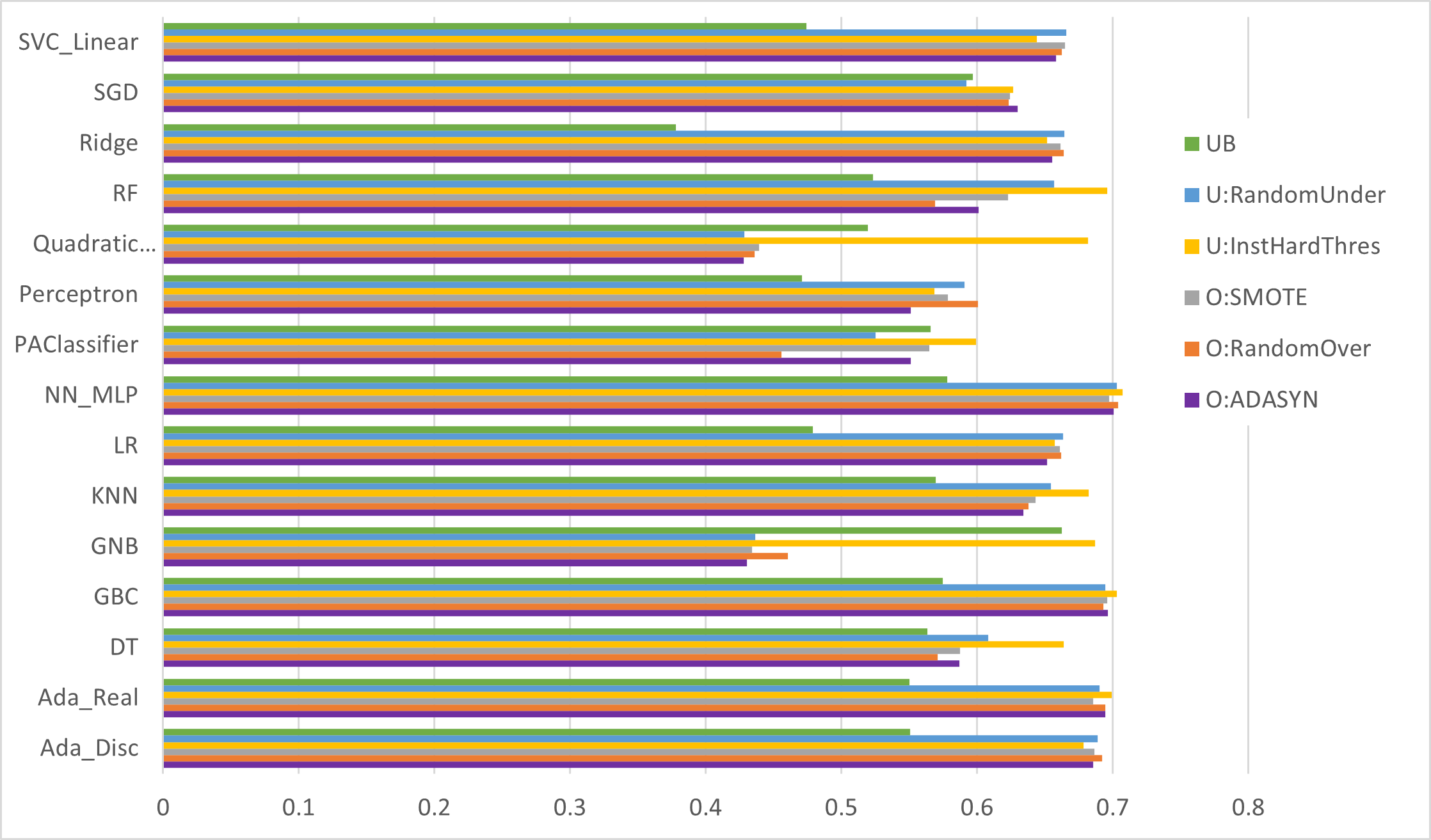}
    \caption{G-mean score - Effect of balancing - secondary data set}
    \label{fig_Gmeans_sec}
\end{figure}
The secondary data, which we PCA encoded to match our primary data set, also confirms the findings regarding the use of undersampling over oversampling as seen in Figure~\ref{fig_F1_sec} and Figure~\ref{fig_Gmeans_sec}. While the two data sets are dissimilar in size, dimensions and domain, they are both unbalanced and PCA obfuscated. However, note that that primary data set has an unbalanced minority fraud percentage of 0.172\% while the secondary data set has 22\% minority frauds.

\subsection{Classifier selection strategy}
Based on the results, we propose the use of a combination of dimensions selection with an intelligent sampling and adopt a data-driven model selection, unique to the data set.

While it was observed that undersampling is generally the best option to use in case of massively imbalanced credit card fraud detection scenarios, \textit{purely} random under sampling does not help; some intelligent algorithm at the balancing stage is needed to effectively increase the final classifier performance. 
It seems that the final classifier could do a better classification job with lesser actual data than in the presence of data that was artificially added by an oversampling process. This is important as focusing on undersampling, particularly in massively imbalanced data set with huge volumes, means that it becomes possible to early-drop larger portions of confident normal classes. This would enable use of larger data sets, as we drop the more common confident normals, allowing more of important fraud classes to arrive at the final classifier. It should also result in a huge resource reduction.

Alternatively, the observation that random undersampling tended to degrade performance for these characteristic data sets reinforces the assumption that rather than re-balancing, a form of intelligent assistance can boost the performance scores.

While PCA does not significantly degrade performance, care should be taken to use the appropriate principle component size to avoid overfitting.

Despite the accepted benefits of ensembles to improve on the constituents, experiments have indicated that there may be certain satisfying conditions to be met for this to take place.

Based on our observations, we propose a classifier selection strategy that uses a combination of feature selection and an intelligent undersampling option. Multiple classifiers should be used on a data set to determine the most appropriate choice. This also collaborates with \cite{dal2017credit} approach to use the data itself to determine the final classifiers. As the data set is unknown, and possibly PCA encoded, feature engineering or data preparation is not suggested and can be avoided with proper dimension selection. Use of ensemble as the final stage is optional and depends on the net gain the ensemble is able to provide. 

\section{Limitations and future works}
According to \cite{dal2017credit,dal2015calibrating}, concept drift where transactions might change their statistical properties over time can be a major characteristic of credit card fraud detection. This requires a constantly learning and updating model, while discarding old leanings and acquiring new leanings; essentially a moving window approximation. Due to unavailability of data, this aspect is not included in this work. 

One of the main issues of the model selection strategy is performance. There is a trade-off between speed and memory consumption. More intermediate results that can be store (preferably in-memory) would speed up the process on subsequent nodes of the data flow we propose in Figure~\ref{fig_frameworkDataflow}. Early dropping of poorly performing models is also a promising area. Essentially, any optimization that allows obtaining the top k, with even part of the data set, should help enhance the performance. Once the K-most models are known, we can use the full data set. 
However, we cannot simply use a smaller subset reliably when using massively unbalanced data for which this framework is designed, without some explicit algorithm. While we have included the training time, more comprehensive measures of performance is needed based on constraints of the model deployment context and environment.

It was observed that ensemble model did not match or outperform the individual constituent component models as expected. Further research could validate this and look into possible explanations. This may be due to improper ensemble construction leading to unfavorable conditions, arising from imbalance, and a worthwhile area to investigate considering the growing popularity of ensemble models. 

\section{Comparison with other work}
Evaluating one-class adversarial net models on the primary data set, \cite{zheng2019one} reports F1 scores of 0.8416 ± 0.0330 for unmodified data and 0.8656 ± 0.0220 for data pre-processed with a LSTM-autoencoder. Our best model, RF with InstanceHardness threshold based undersampling (Table 4, row 1) reported 0.8466, which is better than raw results of \cite{zheng2019one}.

Also, \cite{zheng2019one} used an altered distribution of 700 normals and 490 frauds and negated the effect of massive imbalance. It should be noted that, by using an altered test distribution, classifier scores can move upwards as probability of false-negative misclassifications due to imbalance is almost non-existent, specially at 33\% against 0.2\% in the unmodified data. Scores obtained by our work is more realistic as it reflects the real data distribution.
While there are few other works using the same primary data set, the evaluation metrics used by these researchers are different and does not allow for comparison. Furthermore, our scope is proposing a data driver selection strategy, rather than a model design.

\section{Conclusion}
In this work we provide a classifier selection strategy and framework for massively imbalanced sensitive, encoded data; typical of fraud detection tasks. Specifically, we show that selecting the optimal classifier based on performance of all combinations of PCA feature selection, under sampling and multiple classifiers to be the best option, followed by building a voting classifier to see if that can further enhance the score. This is essentially a data-driven classifier selection. The model that was derived with the strategy outperforms even advanced generative models, whist using the whole massively unbalanced real data distribution validating the effectiveness of the strategy. 

\section{Acknowledgments}
Dedicated to Sugandi.

\bibliographystyle{splncs04}
\bibliography{main}
\end{document}